# CowScreeningDB: A public benchmark dataset for lameness detection in dairy cows

Shahid Ismail, Moises Diaz, Cristina Carmona-Duarte, Jose Manuel Vilar, Miguel A. Ferrer

## Abstract

Lameness is one of the costliest pathological problems affecting dairy animals. It is usually assessed by trained veterinary clinicians who observe features such as gait symmetry or gait parameters as step counts in real-time. With the development of artificial intelligence, various modular systems have been proposed to minimize subjectivity in lameness assessment. However, the major limitation in their development is the unavailability of a public dataset which is currently either commercial or privately held.

To tackle this limitation, we have introduced CowScreeningDB which was created using sensory data. This dataset was sourced from 43 cows at a dairy located in Gran Canaria, Spain. It consists of a multi-sensor dataset built on data collected using an Apple Watch 6 during the normal daily routine of a dairy cow. Thanks to the collection environment, sampling technique, information regarding the sensors, the applications used for data conversion and storage make the dataset a transparent one. This transparency of data can thus be used for further development of techniques for lameness detection for dairy cows which can be objectively compared. Aside from the public sharing of the dataset, we have also shared a machine-learning technique which classifies the caws in healthy and lame by using the raw sensory data. Hence validating the major objective which is to establish the relationship between sensor data and lameness.

## IMPLICATIONS

Lameness detection is one of the main tasks in dairy systems given its implications in the production ambit. However, the data used during detection is generally either held privately or sold commercially. In this study, we create a multi-sensor dataset (CowScreeningDB) which can be used for lameness. As we have made the dataset public[1] and free of charge for researching purposes, it should act as a benchmark allowing to objectively compare the techniques put forth for lameness. We also provide details of the sampling system used, comprised of hardware and a baseline classification algorithm.

## 1. INTRODUCTION

Productivity in livestock farming is negatively affected by housekeeping costs and dairy diseases such as lameness. In recent years, sensor-based artificial intelligence systems have been used to assess the overall health of cows, including behavioral changes, body part detection (Jiang et al., 2019a) etc. Among the physiological behaviors i.e. physical motion of the cow under observation, observed in this context, lameness is the main cause of the most critical change observed in the gait or stance of an animal. Lameness is a collective term for three types of abnormalities namely claw horn disruption lesions (CHDL), skin lesions, and non-foot lameness (Mason Colin, 2007). CHDL is not a single disease, but it reflects a number of non-infectious foot lesions including sole ulcers sole hemorrhage, and white line disease (Griffiths, Bethany E., et al, 2020). Skin lesions are indicative of dysfunctional housing of dairy cows (Kielland, C., et al. (2009) and non-foot lameness generally includes all non-foot lameness like lameness due to bone injury, muscle and joint damage (Mason Colin, 2007).

As lameness is associated with a broad spectrum of health disorders, it can act as a good biomarker to detect these disorders. Therefore, veterinary clinicians are trained for the said task at individual and herd levels. However, lameness detection can be hard to detect and additional interventions like rain need to be considered (Thompson, Alexander John, et al., 2019). Generally, a veterinary clinician can detect lameness by focusing on lameness predictors which are mainly gait and posture traits. Asymmetric gait, reluctance to bear weight, short steps, joint flexibility, arched back, head bobbing, walking speed, difficulty in turning, and difficulty in rising used are predictors which clinician generally uses during a visual inspection (Schlageter-Tello, Andrés, et al., 2014). Recent years have seen increased automation of lameness detection through the use of sensor-based systems that record force distribution, kinematics (linear, temporal and/or angular) measurements or any other combination related to the stance or motion of the animal.

Force platforms are generally used to perform kinetic assessments to detect lameness in dairy animals objectively. These devices have been used by researchers such as Chapinal and Tucker (2012), who employed these platforms along with video recording, for assessing the weight distribution among limbs and step counts in lame cows. In the same vein, Hertem et al. (2013) detected lameness using a tri-variable model using logistic regression, which is based on milk yield, rumination, and neck activity. Another study, which used logistic regression, was conducted by Kamphuis et al. (2013). They used uni-variables such as weight, activity (average steps per hour) and features related to milk, and their combinations which converted analysis to multivariable to assess lameness. Bruijnis et al. (2010) related foot disorder with lameness using a dynamic stochastic simulation model in a similar study. Similarly, Pastell et al. (2006) used piezo electric force sensors for lameness detection.

Kinematics parameters such as stride length, support time, and articular range of motion are the characteristics used by veterinarians to assess lameness. In this context, Poursaberi et al. (2010) introduced a real-time system for lameness detection based on shape analysis using back posture. The authors concluded that posture-based classification is able to operate in real-time. Piette et al. (2020) also employed back posture analysis to validate the performance of an automatic camera-based system. Hartem et al. (2016) compared the performance of a graphic video system for lameness detection with a multi-sensor system composed of milk production, activity, and postural changes. They concluded that sensor-based videography enhances lameness detection on the systems with other kinematic sensors in place. In a related work, Zhaoa et al. (2018) used a multivariable analysis looking at gait asymmetry, speed, tracking up, stance time, stride length, and tenderness to classify a cow as sound or lame. Other studies which used the kinematics were conducted by Vázquez Diosdado et al. (2018); Van De Gucht et al. (2017); Maertens et al. (2011) and Pastell et al. (2008). However, studies by Van De Gucht et al. (2017) and Pastell et al. (2008) are different from the other studies as they are utilizing force sensor based kinematics.

Video and image based lameness detections reflect the major modalities utilized in the fields of automated lameness detection systems as lameness can be detected online as well offline using these systems. Among studies considered are the ones conducted by Jiang et al. (2022), Kang et al. (2020), Jiang et al. (2020), Piette et al. (2020), Jiang et al.,-b (2019), Zhao et al. (2018) and Van Hertem et al. (2014). Unlike digital systems which are utilized for lameness detection (Kang et al. (2020); Jiang et al. (2020); Piette et al. (2020); Jiang et al.,-b (2019); Zhao et al. (2018) and Van Hertem et al. (2014)),  Jiang et al. (2022) used an analog system ( phase alteration line (PAL)). Due to high financial and computational cost, these systems are generally limited in storage. For example, Jiang et al. (2022), Kang et al. (2020), Jiang et al.,-b (2019) and Zhao et al. (2018), are generally limited to 40 seconds, 1k images, 40 seconds, 30 seconds and 7 minutes, respectively. The systems by Jiang et al. (2022) and Jiang et al.,-b (2019) are the real time systems which used using back position of the cows and double normal distribution statistical model, respectively. However, classification of lameness by Kang et al. (2020), Piette et al. (2020), Zhao et al. (2018) and Van Hertem et al. (2014) is offline in nature and they are utilizing supporting phase, back posture, leg swing analysis and consecutive night-time milking sessions, respectively.

To reduce the cost associated, inertial sensors such as pedometers and accelerometers have gained in popularity in the recent past. They are in wide spread use as reflected by studies like Lemmens et al. (2023), Frondelius et al. (2022), Borghart et al. (2021), Jarchi et al. (2021), Antanaitis et al. (2021), Shahinfar et al. (2021), Taneja et al. (2020), Byabazaire et al. (2019), Weigele et al. (2018), Barker et al. (2018), Vázquez Diosdado et al. (2018), Beer et al. (2016), Thorup et al. (2016), Thorup et al. (2015), Garcia et al. (2014), De Mol et al. (2013), Van Hertem et al. (2013), Kamphuis et al. (2013), Van Hertem et al. (2013), Chapinal and Tucker (2012), Maertens et al. (2011) and Nielsen et al. (2010).

The main reason for the popularity of inertial systems is the continuous sampling of lameness predictor.

Generally, lameness is detected using inertial measurements (Borghart et al., 2021; Jarchi et al., 2021; Weigele et al., 2018; Barker et al., 2018 and others), milk related measurements (Lemmens et al., 2023; Borghart et al., 2021; Van Hertem et al., 2016 and others ), behavior related predictors like lying time, number of lying bouts, maximum length of the lying bout, roughage feeding time etc. (Frondelius et al., 2022; Zhao et al., 2018; Thorup et al., 2016 and others ) and the mixture of the mentioned predictors (Riaboff et al, 2021; Shahinfar et al, 2021 and others).

Aside from modalities just mentioned, videography for sensor data represent a composite approach where computational load of videography is reduced by using classification based on sensor data. Among these studies are the research work conducted by authors like Beer et al. (2016), Van Hertem et al. (2016) and Kokin et al. (2014). A summary overview of the all the modalities just presented in shown in the literature review which is given in Table A (Supplementary material).

In the table A, studies are compared using hardware (sensor) used to sample the prediction parameter (signal), data statistics, lameness levels and public sharing. Data statistics are given by Equations 1 and 2.

$$S_S = N_C \left(N_S * N_{D/O}\right) \quad (1)$$

$$S_V = N_C \left(N_S * I_R * D\right) \quad (2)$$

Where, $S_S, S_V$ are sensor and video/image based statistics, respectively. $N_C, N_S, N_{D/O}, I_R$ $and$ $D$ are number of cows, sensors/video (images) files, number of days/observations, image resolution in pixels and duration for videos, respectively. Here, $S_S$ reflects two types of sensor based results. In first kind, results are reported in form of number of cows, number of sensors, duration for which sensor was active. Results are also reported in simplified form of number of cows, number of sensors and number of observations. Results are differentiated by * which reflects the second kind. Similarly, $S_V$ reflects both video as well as image based results reflected by ! sign. Here, distinction is made between these adding two! sign for video based research works. Again, the results are given in terms of number of cows, number of images/video files and duration, respectively.

An objective comparison can be done in terms of number of observations/ files from a given number of cows. This comparison is valid for image/video based studies as well as for the sensor based studies where the number of observations/files are given. Jiang et al. (2022), Kang et al. (2020), Zhao et al. (2018), Lemmens et al. (2023), Borghart et al. (2021) are such example studies which carry necessary information for comparison. However, this comparison is not valid for Frondelius et al. (2022), Antanaitis et al. (2021) and similar studies. Further complication is added in videography based studies when duration of videos is considered in comparison to images. Similar difficulty also arrives in sensor based studies which are based on samples (Shahinfar et al., 2021; Borghart et al., 2021) verses studies containing duration of sampling (Riaboff et al, 2021; Byabazaire et al.,2019). The information just mentioned highlights a critical challenge in the objective comparison of studies for lameness detection. A further complication is objective comparison in similar studies. For examples, it is a challenge to compare Borghart et al., 2021with Jarchi et al., 2021. Both studies are reported in terms of cows, number of sensors and the observations/files. However, we have 3799 observations from 164 cows using 6 sensors in contrast to 25624 observations from 23 cows using only eight sensors. A further layer of difficulty is added by the lameness scoring. Generally scoring is reported in terms of Sprecher et al. (1997) which is from 1-5. However, Frondelius et al. (2022) and Jarchi et al. (2021) have used the pairs of scores for lameness levels. Similarly, certain scores are missing in studies (Garcia et al. (2014) and Barker et al. (2018)). The hardware used is very diverse as reflected by the use of RumiWatch noseband halter, SONY HDR-CX290E, Bosch BMI160 Inertial measurement unite etc. Finally, all studies except the present study have not granted public access to its data, limiting the ability to do an objective comparison of the developed techniques.

All the complicating factors just mentioned for objective comparison point to creation of a public dataset on which a technique could be objectively validated. In order to remove this major limitation, we have put forth, CowScreeningDB[1], along with the necessary information. Critical importance of the data in the domain of machine learning cannot be stressed more as reflected by the research works (Martens, B. (2018; Zhao, et al. (2020); Celi, A., et al. (2019); Artrith, N., et al. (2021); Shimron, et al. (2022); Rodgers, et al. (2023); Aldoseri, et al. (2023); Jain, et al. (2020); Hettinga, S., et al.(2023); Hu, W., et al. (2020); Catillo, M., et al. (2022); Dekker, Ronald. (2006); Paullada, A., et al. (2021) and Trisovic, A., et al. (2021)) to name a few ones. Among studies just mentioned, research endeavors (Zhao, et al. (2020); Hettinga, S., et al. (2023); Hu, W., et al. (2020); Catillo, M., et al. (2022) and Celi, A., et al. (2019)) reflect the importance of public access of data for machine leaning based AI development. Aside from public access, the quality of dataset must be ensured so that it could be reused. Infect, data collectors are expected to follow the FAIR (Findability, Accessibility, Interoperability, Reusability) principal (Martin, E., et al. (2017); Roche, D., et al. (2022) and Llebot, C. & Steven V. T. (2019)) policy for public sharing. Among public datasets following FAIR are the datasets from digital signal processing group (GPDS, 1992), Center of Excellence for Document Analysis and Recognition (CEDAR, 1991), Biometrics and Data Pattern Analytics (BiDA, 2018) etc. Similarly, CowScreeningDB, is also collected following the directions just mentioned and is now shared at multiple public repositories[1].

Authors like to point out that private datasets from inertial equipment similar to our study are available. For example, Benaissa, S., et al. (2019) have collected data using accelerometer but during study the dairy cows stayed in the barn, and the sampling frequency is less than 1 Hz. In this way, it is not possible to compare their results with our database, in which the dairy cows move from the barn to the milking parlor, and the sampling frequency is over 100 Hz. Hence, to the best of our knowledge, there is not an equivalent dataset for comparison. The present study is the only one freely sharing full inertial measurement unit raw data (accelerometer, gyroscope, and magnetometer) sampled at over 100 Hz, making it the only dataset available for public research and development for lameness detection in dairy cows. Hence, public sharing also represents the main contribution of our study.

The hypothesis behind data collection is to study the relationship between sensory data and lameness scores. The study establishes the mentioned relationship using a machine learning based inference technique while utilizing the classical classification measures, hence, fulfilling the major objective. In short, we are not only offering a database but we are aslo providing a baseline technique along with the database. It will help the researchers to continue working on this research.

The following are the major highlights of our research:

- A labeled dataset, CowScreeningDB, is made available to public. This dataset is collected from both healthy and lame cows; the lame cows are characterized by different lameness levels, which were assigned by trained veterinary clinicians. The reference provided along with the dataset allows an unbiased assessment of lameness.
- During CowScreeningDB sampling, each cow is observed for approximately 6.7 hours during routine life. The continuous sampling as the dataset is created ensures that it is transparent, and can thus be used for continuous evaluation.
- Information regarding the sampling and classification system are also shared, allowing its reuse by other researchers or domain experts.
- A benchmark containing standard measures is introduced, allowing to objectively compare different techniques.

## 2. MATERIAL AND METHODS

The material and methods section is divided into five major subsections: data collection, data distribution, benchmarking criterion, and methodology introduced, which includes features used as part of the methodology.

### *2.1.* ***Data collection***

Farm details and data sampling are included in data collection. First details of the farms which are utilized during the study are given below.

#### ***2.1.1 Farm details***

The biometric data for this study were obtained from an intensive dual-purpose Friesian dairy cattle farm situated in Agüimes, Las Palmas, Spain. This farm, boasting a total of 1100 animals, focuses on milk production, generating an average of 5,750,000 liters annually. Each cow contributes to this figure by producing an average of 12,700 liters per year, equivalent to 35 liters of milk per day. The herd consists of 580 lactating cows, 220 cows in rebreeding, and 300 designated for meat consumption. The lactating cows, aged between 1 to 9 lactations, are divided into different pens based on production days and capacity, whether primiparous or multiparous. The farm employs a strategic approach in managing the animals, ensuring their well-being and optimal productivity.

To maintain a representative sample, both primiparous and multiparous animals were included in the study. Care was taken to select both healthy and pathological animals from different pens to avoid bias in the results. The animals' gait and pathology were assessed using a 5-point locomotion scoring system developed by Sprecher et al. (1997), focusing on back arching, weight distribution on limbs when standing, and movement patterns. The farm's facilities lack individual cubicles, ensuring a minimum of 10 square meters of space per animal. The farm has implemented an artificial insemination program for over a decade, enhancing genetic diversity and overall livestock quality. The corral bedding, a mixture of manure and straw, is regularly oxygenated to prevent bacterial growth. Concrete surfaces in feeding and milking areas facilitate easy cleaning.

For the study, a sample size of 25 to 30 animals, both healthy and pathological, was determined. This number allowed for comprehensive data collection within a reasonable timeframe. Recordings, lasting 5 to 6 hours, captured the animals' natural movements. In specific cases, where animals were in the drying phase and didn't need to travel to the milking parlor, the device recorded for 12 to 24 hours, ensuring thorough movement data. The study accounted for various factors influencing lameness incidence, including traumas, infections, and metabolic disorders, often exacerbated by adverse weather conditions. Regardless of the cause, the focus was on differentiating between healthy and sick animals, enabling prompt detection of issues. Traumas resulting from recent corral adaptations were a notable cause of lameness in this study.

Classification and observation were carried out by a professional veterinarian with six years of extensive experience in dairy cattle. Animals were assessed 1 to 2 times, and rechecks were performed within 12 hours, ensuring consistency in classifications. Previous studies (Eriksson, Hanna K., et al., 2020; Sahar, Mohammad W., et al., 2022) corroborated the reliability of this approach, affirming the stability of classifications even after 24 hours.

### *2.1.2 Data Sampling*

Data sampling and transfer are carried out using network diagram as shown in Figures 1-3. From network diagram (Figure 1), we can see that dairy cows are fitted with smart watches which also have wifi connectivity. These watches are connected with a smart phone which intern is connected with a cloud service. From cloud service, data is given as an input to the classification algorithm which classifies cow as either healthy or lame. In Figure 2, details regarding the sampled file are given. From the Figure, we can see that file was saved using comma separated values (CSV) on 15 May 2022. File name also include the start time of storage along with the position of sensor in this case. To record the full database, a smart watch was located on one of the cow's four legs as can be seen in Figure 3. In Figure 3, an arrow 'C' reflects the general location of the sensor on the back side of the watch (Apple Watch 6, Apple).

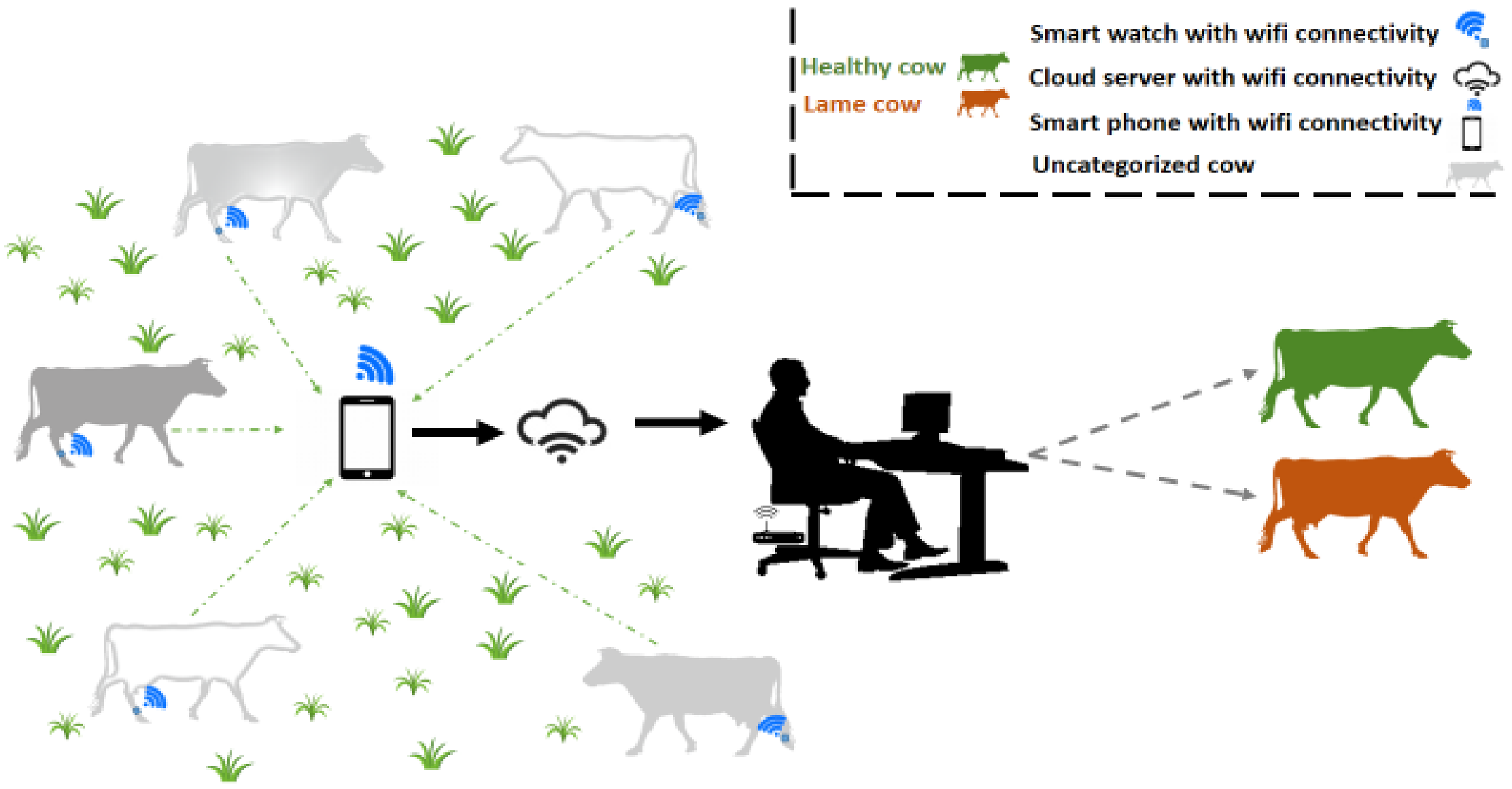

***Figure 1:*** *Network Diagram. Figure shows the main components utilized in the sampling of data and classification. Smart watches are attached in the lower part of hind legs which are connected with smart phone. An application is installed in the smart phone which converts raw data collected from smart watches to format suitable for saving on cloud serves. Classifier who is generally a person downloads the data from cloud and feed it to machine learning based classification algorithm which categorizes cows in terms of healthy/lame cow.*

Our teams at the University of Las Palmas de Gran Canaria created an App specifically for capturing data in this study. The app developed for data acquisition allows the veterinarian to select the paw where the watch will be located. As such, the registered leg, recording date, and animal identification number are included in each stored file name. The watch is recording continuously and the data are saved in consecutive 90-second files which is then saved in consecutive 90-second files. Once the data is recorded on the smart watch, it is synchronized with a smart phone and uploaded to cloud for storage. Each file contains 13 columns with temporal information data, cow-generated acceleration (without gravity) for all three axes of the device, gravity (three axes), gyroscope data (radial velocity) of the three axes, and the attitude (yaw, pitch, and roll). Although Apple Watch 6, Apple iphone and iCloud were used for smart watch, smart phone and cloud service, respectively but any other smart watch, smart phone and cloud service combination can also be utilized.

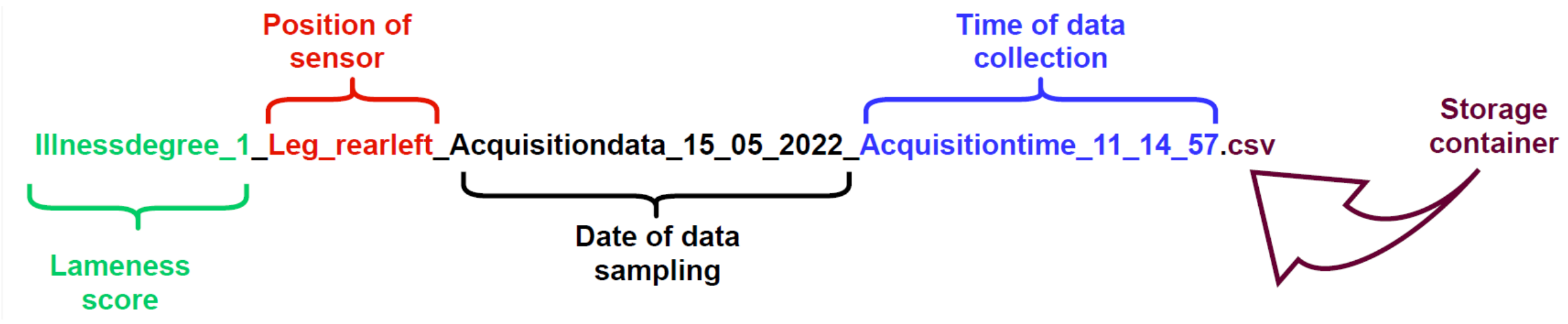

***Figure 2:*** *Figure shows the name of a file which itself contains the sampling information. From Figure, it can be seen that this specific file was sampled from a cow which was healthy as its lameness level was 1. The location of sensors is on the rear left leg and data was collected at 14 minute and 57 seconds past 11 clock on 15 May 2022. The data was saved using the comma separated value based file format system.*

Three sensors, namely, a gyroscope, a magnetometer and accelerometers, were used in the study. Information regarding acceleration and radial velocity was collected using the gyroscope and accelerometer directly. Similarly, the Apple watch also provided information regarding gravity post- processing. However, for information on the attitude (radial position), the direction of the reference pole using a magnetometer was also initialized. The sampling rate, i.e., the sampling frequency used, was 100 Hz. Afterwards, the data was transmitted to the mobile phone and the cloud service. Data can be appended in serial or parallel, giving 43 cows and 11,518 samples, respectively. In Section 2.2, the statistics for, CowScreeningDB, are given. An analytical overview of the multi-sensor data is given in Table 1. From Table, it can be seen that data is composed of different physical signals i.e. acceleration, gravity, radial velocity and attitude. Hence, data is separable when different signals are considered separately.

***Table 1**: Details of the signal in the datasets. Where, ST is signal type for which respectively units are also given. Acceleration, gravity, angular rotation and roll, pitch, and yaw all have three components which are directed along x, y and z axis.*

| Channel No./ ST(Unit) | Time | Acceleration | Gravity | Angular Rotation | Roll, Pitch and Yaw |
|---|---|---|---|---|---|
| 1 | S | - | - | - | - |
| 2:4 | - | $m/s^2$ | - | - | - |
| 5:7 | - | - | $m/s^2$ | - | - |
| 8:10 | - | - | - | Radian/s | - |
| 11:13 | - | - | - | - | Radian |

The device placement on limbs was random and it was even placed on the affected limb as well. Random placement of sensor helped in eliminating bias and proving the sensor's capability to detect lameness without targeting specific limbs. Animals adapted well to wearing the lightweight sensors, and initial minutes of recordings, during placement manipulation, were excluded to ensure accurate data collection.

### *2.2 Data distribution*

***Figures 4, 5*** show the data collected during CowScreeningDB and the necessary statistics related with it. From ***Figure 4,*** it can be seen that there are five categories of cows in terms of lameness. A cow with lameness score of 1 represents a sound (healthy) cow followed by cows at lameness scores of 2-5 which represent cows at various levels of lameness. Figure 4-a also shows binary categorization of cows in terms of a healthy or lame cow. The number of healthy are 19 and lame cows at lameness scores 2-5 are 7, 6, 6 and 5, respectively. In Figure 4-b, the number of samples per lameness score is presented. There are 4787 samples (a file of 90s) taken from healthy cows which are 19 in numbers as reflected in Figure 4-a. Similarly, 1038, 1630, 2432 and 1631 samples are respectively collected from lame cows. As shown, there are 19 verses 24 cows for the healthy and lame categories with 4787 and 6731 samples, respectively.

In ***Figure 5***, data is presented in form of average number of samples and average time duration for the different lameness scores.  From the boxplots shown, it can be inferred that the data varies between different lameness scores both in terms of samples as well as time duration.

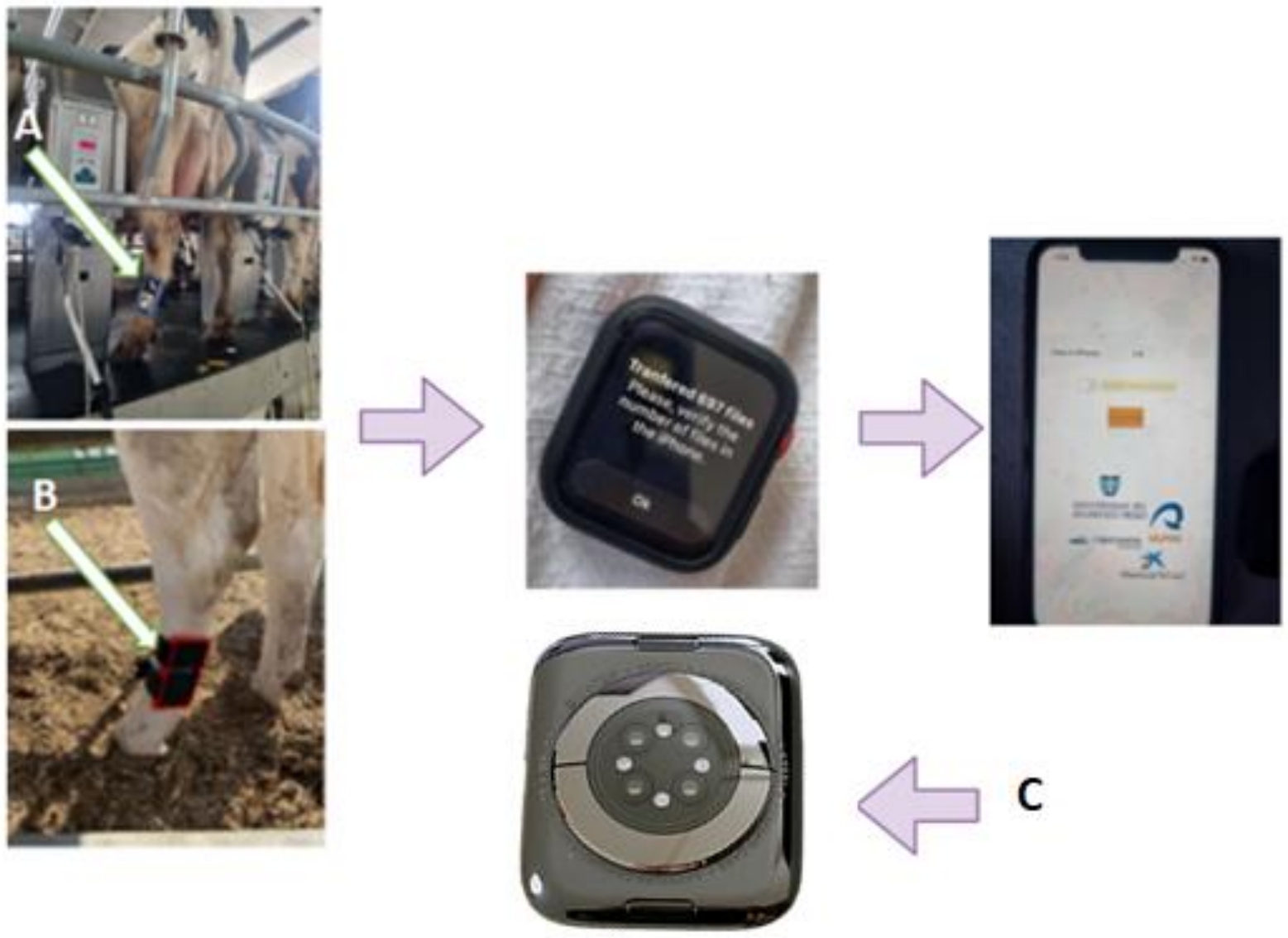

***Figure 3****: Data transfer system is composed of sensor attached to the cow, an Apple watch 6 and an iPhone App. An Apple watch was recording data during the routine life of a cow. Arrows 'A' and 'B' show the watches attached to the paw of the cow during two dairy activities: during milking in milking parlor and a normal walk. An arrow 'C' points to the general location of the sensors inside the watch (Apple Watch 6, Apple). Once data was collected, they were transmitted to a local system as well as to a cloud service.*

### *2.3 Benchmarking criteria*

For CowScreeningDB, the benchmarking criteria for the evaluation of technique are given below.

- Classification benchmark: *Table 2* indicates that data could be used for binary as well as multi-class classification using the standard measures of sensitivity, specificity, precision and accuracy. If the binary class points towards an abnormal animal then the multi-class classification screens the particular level of lameness for that case.
- Ablation study: An ablation study is generally conducted to select the most discriminating features or signals.

Since four different types of signals were used in this study, any novel technique introduced therefore will have to report results regarding the most discriminating signals and features used. Studies could be further extended to include the fusion of different signals like usage of acceleration along with gravity and other such combination.

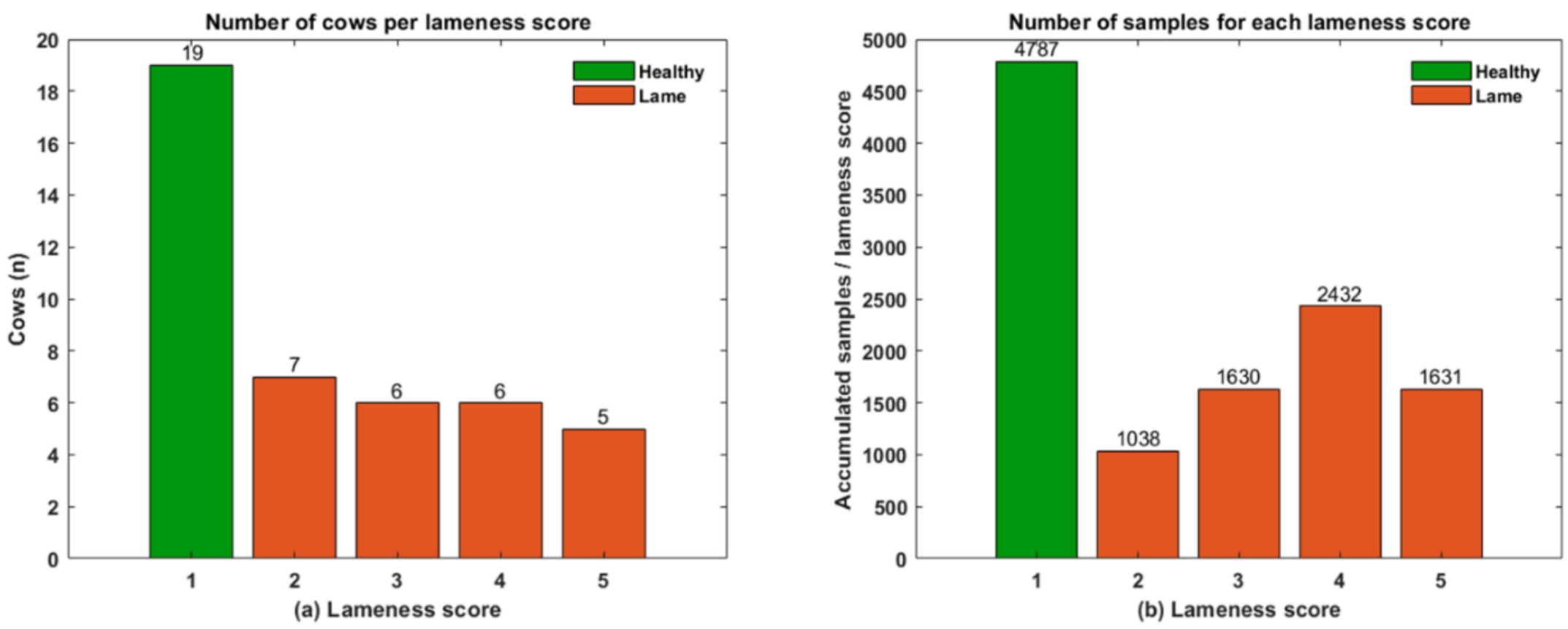

***Figure 4**: Data statistics for CowScreeningDB using lameness scores. (a) The figure shows the data division in terms of designated lameness scores (1-5), with level 1 reflecting the healthy cow. Lame cows are represented by lameness scores of 2-5. (b) The figure shows the data in terms of the number of samples/lameness score.*

The minimum criterion for benchmarking is the binary classification, as mentioned above, along with an ablation study regarding the selection of the most discriminating signals and features. However, an in-depth study could be extended to include multi-class classification. In the next section (Section 2.4), the details of a machine learning-based classification system are given, and could be used as a baseline for future studies.

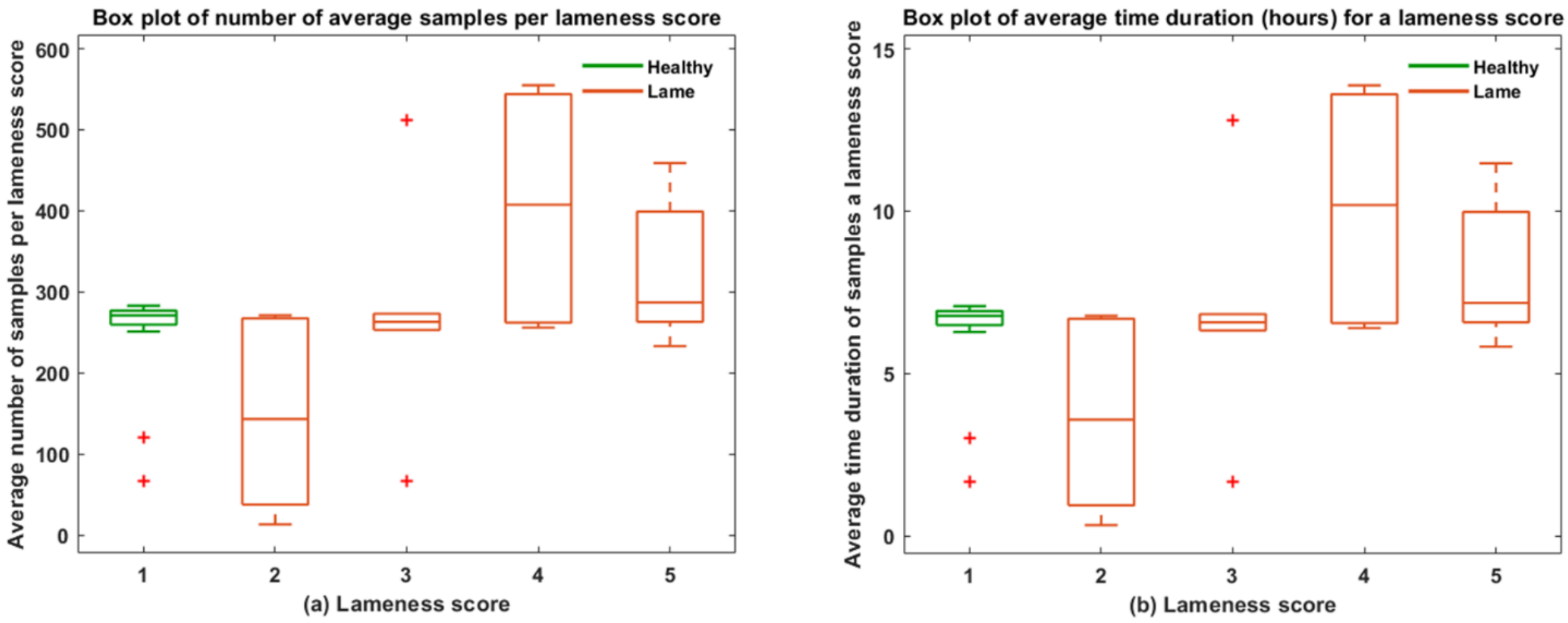

***Figure 5:** Data distribution in terms of average number of samples and average time duration related with these samples. (a) Box plot of average number of samples for different lameness scores (b) Box plot of average time duration of samples collected for a lameness score. The average time duration is in hours.*

### 2.4 *Methodology*

Once data is converted into the serial format, it is then, classified using the machine learning-based classification system. However, data conversion to the serial format brings a challenge in terms of length variation among the samples. In order to address the length variation, a length normalization could be done. In present case, length variation among signals makes them unsuitable for length equalization using normalization method due to large variation in length. Therefore, features which could reflect the global nature of signals are extracted. For example, power present at different frequencies in frequency spectra could reflect the frequency behavior related with inertial data. For present study, feature extraction is applied for segmented as well as non-segmented signals as given in classification technique in Figure 6. The system can be divided into three main modules, namely, segmentation algorithm, feature calculation, and classification network. The modules are introduced below, starting with the segmentation algorithm.

**Segmentation algorithm:** Comprises $3^{rd}$-order moving median filter, $8^{th}$-order zero phase filter Homomorphic filter with cut off of 3 Hz, and normalization. A moving median filter removes the spike noise from the signal, which is then given as an input to the Homomorphic filter. Homomorphic filtering uses the Hilbert transform to enhance the average components. The details of operations applied during Homomorphic filtering are given in Equations 3-7 and *Figure 7*. From Equations, we can infer that during Homomorphic filtering, signal $x[n]$ is considered as being made up of low, $f[n]$, and high frequency, $h[n]$, component with low frequency, $f[n]$, referring to the lameness contents in present case. To filter these contents, Hilbert transformation along with fileting and log based manipulation are applied. Figure 7 shows the sequence in which these operation are applied and the operation at the start is Hilbert transformation. The cutoff of the low pass filter is at 3Hz which was set empirically.

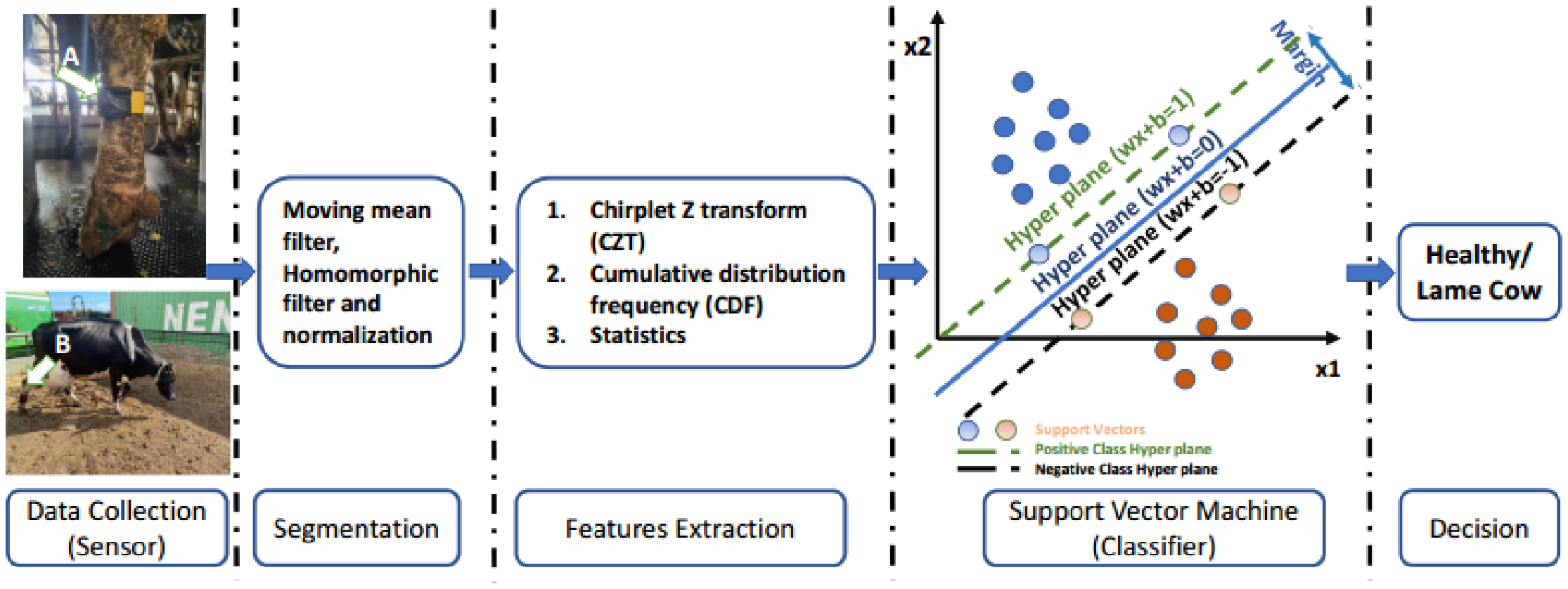

***Figure 6:*** *Classification system is composed of sensor-based data acquisition followed by its conversion into the format suitable for the input to a machine learning algorithm. An SVM (support vector machine) classifies the data in terms of healthy and lame cow. Data is sampled during the routine life of a cow, as shown in the figure. Under "Data Collection: (Sensor), arrows 'A' and 'B' highlight the sensor, i.e., the Apple Watch 6.*

Finally, normalization ([0,1]) is applied to the filtered signal. For the segmentation algorithm, every sample with a power greater than a threshold (10%) is considered as motion, with the remaining samples representing resting positions. Lameness predictors, such as gait symmetry, steps per unit time and ratio of rest to motion are calculated during motion. Hence, segmenting a signal into motion and rest can aid in lameness detection. However, in this study, segmentation is used to increase the number of features (Section 2.4) from 184 to 370 for lameness detection.

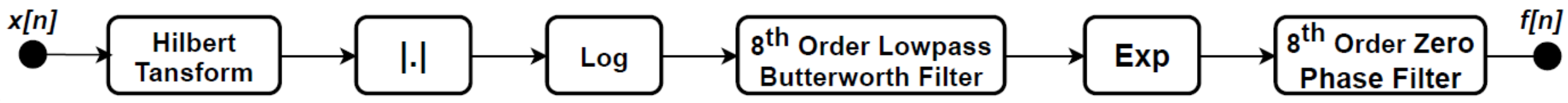

***Figure 7:*** *Figure shows the process of Homomophic filtering. Basic building blocks are Hilbert transform, absolute value calculation, log and antilog transformations and filtering. Here, low pass and zero phase filters are applied to retain the low frequency along with the mentioned processes.*

$$x[n] = h[n]f[n] (3)$$
$$log(x[n]) = log(h[n]f[n]) (4)$$
$$log(x[n]) = log(h[n]) + log(f[n]) (5)$$
$$e^{log(x[n])} = e^{log(f[n])} (6)$$
$$x[n] = f[n] (7)$$

**Features:** Features are characteristics or attributes which can be used to discriminate a healthy cow from a lame one. Two such features, namely, the cumulative distribution frequency (CDF) and the Chirplet Z transform (CZT), are shown in *Figure 8. From Figure 8-a*, it can be seen that the CDF slope for lame cow is close to unity, but it is changing continuously for a healthy cow. A change in slope refers to an abrupt change in power when the cow is in motion. Hence, a healthy cow moves with more power and agility in comparison with a lame one. Similarly, in *Figure 8-b*, the distribution of frequency contents is shown for an acceleration signal. From the figure, we can see that the morphology (shape) of spectral contents can be used to differentiate between a lame and a healthy cow. Features used during the study are primarily based on the mentioned features, along with the power-based features. Details regarding features are given below, starting with transform-based features.

- **Chirplet Z Transform features:** are spectral features that are calculated using the Chirplet transform. The Chirplet Z Transform (Equation 9) is a modified version of the Fast Fourier Transform, which is given below in Equation 8:

$$X(K) = \sum_{n=0}^{N-1} x[n] e^{-\frac{j\pi nk}{N}} \qquad \mathrm{n, K} = 0, \cdots\cdots \mathrm{N} - 1 (8)$$

$$CZT[x(n)] = X(Z_k) = \sum_{n=0}^{N-1} x[n] z_k^{-n}$$

$$z_k = AW_k = A_0 e^{j\theta_0} W_0 e^{-j\varphi_0} \text{ n, K } = 0{,}1 \cdots\cdots \text{N} - 1 \text{ (9)}$$

In these equations (8,9), x[n] is the input signal and X(K) and $X(Z_k)$ are the Fast Fourier Transform and the Chirplet Z Transform, respectively (Hu, Guo-Sheng, and Feng-Feng Zhu, 2011). From Equations (8,9), it can be inferred that the Fourier and Chirplet Transforms are calculated on a unit circle ($e^{-\frac{j\pi nk}{N}}$) and an arc, which is parameterized $(A_0 e^{j\theta_0}, W_0 e^{-j\varphi_0})$. The Chirplet Z Transforms are calculated for original as well as segmented signals. The resultant transforms are both normalized ([0,1]) and greatly decimated with a decimation factor. Three statistical features, namely, the mean, standard deviation and inverse coefficient of variation, are also calculated from the decimated and normalized Chirplet Z transforms.

- **Power-based features:** The motion percentage, the power percentage and the power crossing are global power-based features. The motion percentage is a heuristic feature that is based on the consecutivity of the samples, and is zero for consecutive samples. The power percentage is the ratio of the absolute power of segmented to unsegmented signals (Equation 10) and the power crossing reflects the point where the Chirplet spectrum crosses its mean.

$$PP = \frac{P_s}{P_{us}} (10)$$

In Equation 10, $P_s, P_{us}$represents the power in the segement and unsegmented signal, respectively.

- **Cumulative distribution frequency (CDF)-based features:** are features based on the power profile of the signals. The major steps used to calculate these features are shown in Figure 8-c. Initially, a signal amplitude is converted into an absolute amplitude, which is then converted into a power profile using Equation 11 given below. For this study, the power profile is decimated such that N becomes equal to 90, as shown in Figure 8-a.

$$X(K) = \sum_{n=1}^{K} x[n] \qquad \text{n, K } = 1{,}2, \ldots\ldots \text{N(11)}$$

The total number of features used in the study is 370, as shown in Table 2.

• **Classification network:** The support vector machine is a classifier which uses a decision boundary, support vectors and hyper-planes to distinguish between classes. Support vectors are placed on the positive (Class 1) as well as negative hyper-planes (Class 2). The support vector machine tries to increase the distance between both hyper-planes in order to distinguish classes [20, 21]. The following is the model used for classification**:**

$$y = \begin{cases} +1, if\ \vec{X}\cdot\vec{\omega}+\vec{b} \geq 0. \\ -1, if\ \vec{X}\cdot\vec{\omega}+\vec{b} < 0. \end{cases} (12)$$

In Equation 12, $X$, w, $b$ and $'\cdot'$ are the input (features), weights, biases and dot operator. As the model in the equation is linear model, w and b represent the slope and y-intercept, respectively. Moreover, the dot product is conducted between the features and weights, and the result is added to the biases. From Equation 12, it is apparent that Equation $\vec{X}\cdot\vec{\omega}+\vec{b} > 0$ points to one class, i.e., a healthy cow, in the present study. A failure of this condition will give label of other class in the case of binary classification. A pictorial representation of a support vector machine is shown in *Figure 6*.

For a binary SVM, a linear SVM with 3rd order polynomial kernel function is utilized along with iterative single data algorithm solver. Default values are used for box constraints (Cost ([0 1]), empirical prior probabilities, nonstandard predictor values and initial weight set to [1 1]) with zero probability for outliers. In the present study, K-fold validation is also applied, which gives binary classification in terms of healthy versus lame cows. The classification system results are given in Section 3. The technique just presented was implemented using MATLAB 2022a on a Dell Inspiron 15 7000 Gaming series computer with 16 Gb RAM, Ci7, 7th Generation, and 4 Gb GPU RAM.

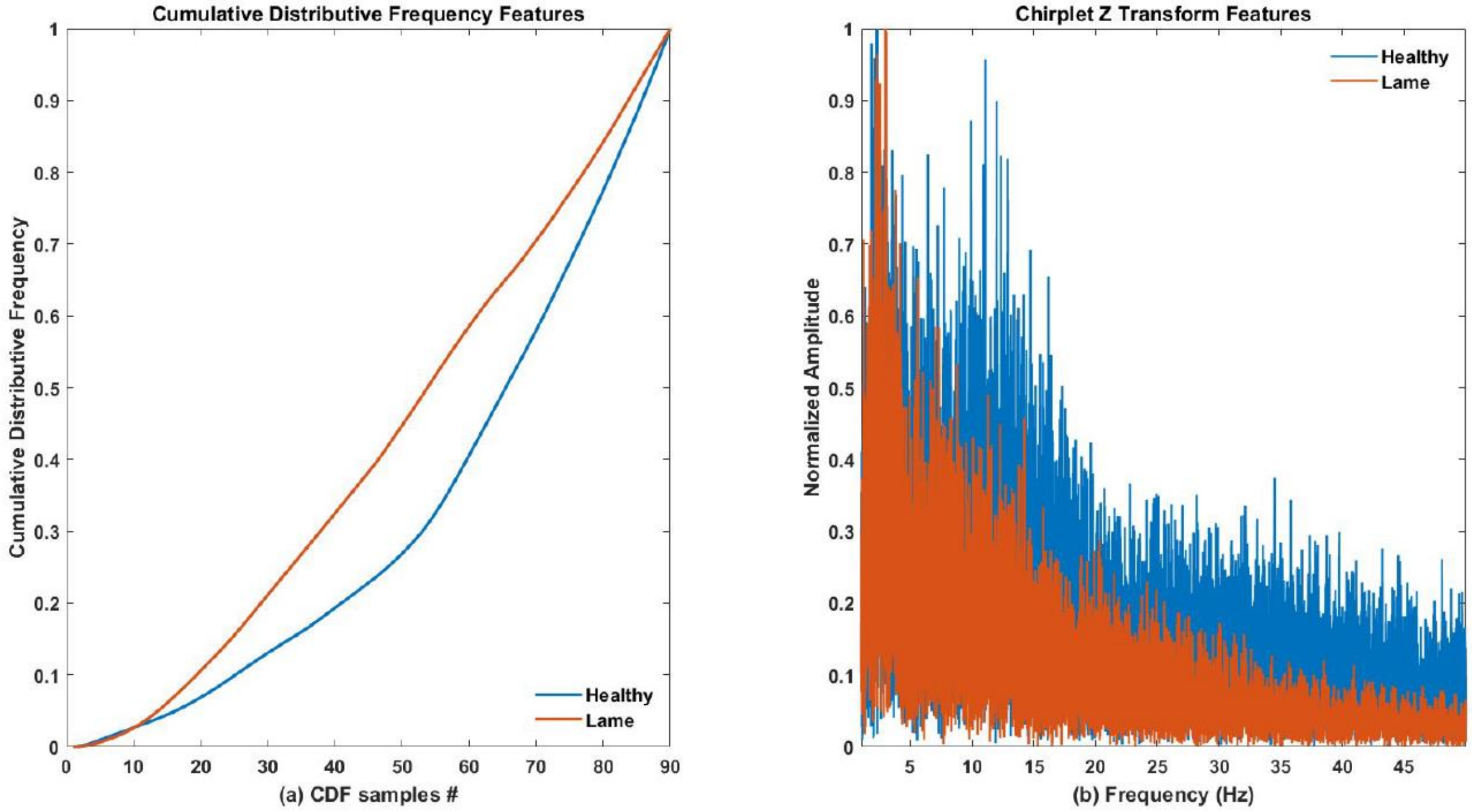

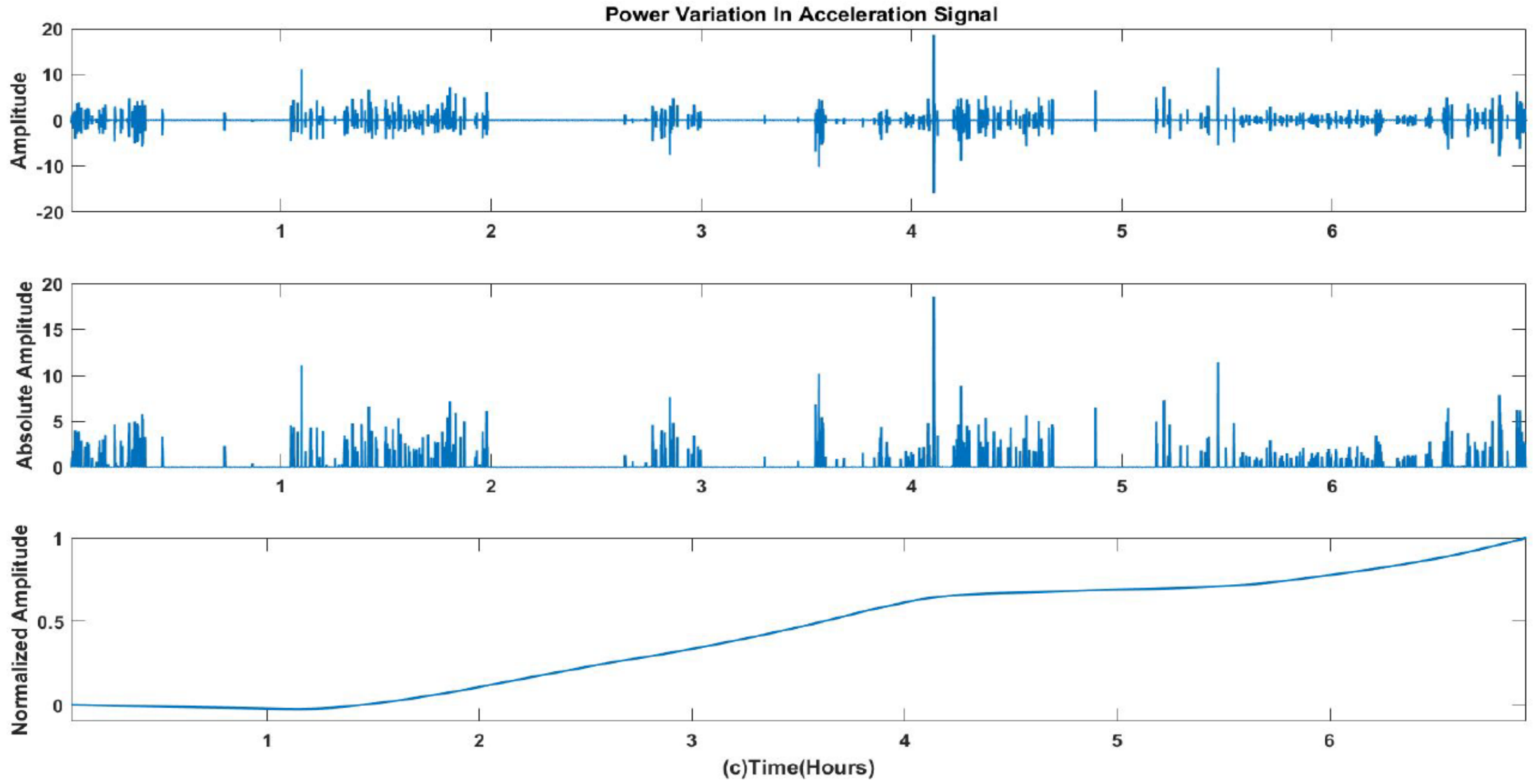

***Figure 8:*** *Features used during the study for classification and procedure to calculate cumulative distribution feature. (a,b) Figures show the cumulative distribution frequency and Chirplet Z transform-based features. From the figures, healthy and lame cows can be discriminated using cumulative distribution feature as the slope of healthy and lame profiles can easily be differentiated. Similarly, the distribution of the spectral contents of the lame cow is different from healthy cow's. (c) First, the signal amplitude is converted into an absolute amplitude. Then, Equation 11 is used to convert the absolute amplitude into cumulative features, during which decimation is also applied.*

## 3. RESULTS

The performance of the presented technique is validated using the minimum criterion given above, i.e., binary classification (healthy/lame cow) along with ablation studies. The criteria applied are named Protocol-I and Protocol-II hereafter, and evaluation metrics used in the study are given below in Equation 5.

***Table 2:*** *Dimensionality and parameters of features used during the study. The total number of features is 370, and out of these, 362 are either CZT- or CDF-based features, where* $\boldsymbol{N_M}$, $\boldsymbol{N_H}$, *CFV,* $\boldsymbol{\delta_{Avg.}}$ *and DF represent the filter orders of moving median, filter orders of Homomorphic filters, cross-fold validation, average threshold, and decimation factor, respectively.*

| $N_M$ | $N_H$ | Normalization | $\delta_{Avg.}$ | DF | CZT | Statistical | Power | CDF | Total |
|---|---|---|---|---|---|---|---|---|---|
| 3 | 8 | [0,1] | 90 | 100 | 2x90 | 2x3 | 2x2 | 2x90 | 370 |

In Equation 5, True positive (TP) and True negative (TN) refer to the situations when a 'healthy cow' is detected as a 'healthy cow' and a 'lame cow' is not detected as a 'healthy cow'. False positive (FP)

$$Sensitivity = \frac{TP}{TP + FN}$$

$$Specificity = \frac{TN}{TN + FP}$$

$$Precision = \frac{TP}{TP + FP}$$

$$Accuracy = \frac{TP + TN}{TP + FN + TN + FP} \quad (13)$$

and false negative (FN) convert a 'lame cow' into a 'healthy cow' and a 'healthy cow' is detected as a 'lame cow', respectively. Using these definitions, evaluation measures of sensitivity, specificity, precision and accuracy are defined in Equation 13. Along with the above measures, an area under curve (AUC) using a receiver operator characteristics (ROC) curve is also used. An ROC curve is a graphical tool to show the performance of a classification model at all classification thresholds. It is plot of a false positive rate (1-specificity) along the x-axis and of a true positive rate, i.e., sensitivity, along the y-axis. As mentioned above, AUC can be calculated using the ROC curve. The larger the area, the greater the classification accuracy of the classifier used.

*Protocol I.* During Protocol I, all features from all signals, i.e., acceleration, gravity, angular position and angular velocity, are used. Hence, all 12 signals and features calculated from these signals are used. *Table 3* and *Figure 9* present the performance of the system. The table shows that the best and worst accuracies are 77% and 46%, respectively. The areas under the curves are 0.69 and 0.38, calculated using ROC curves for the biggest and the smallest areas, and these are reflected in *Figures 9 (a, c).*

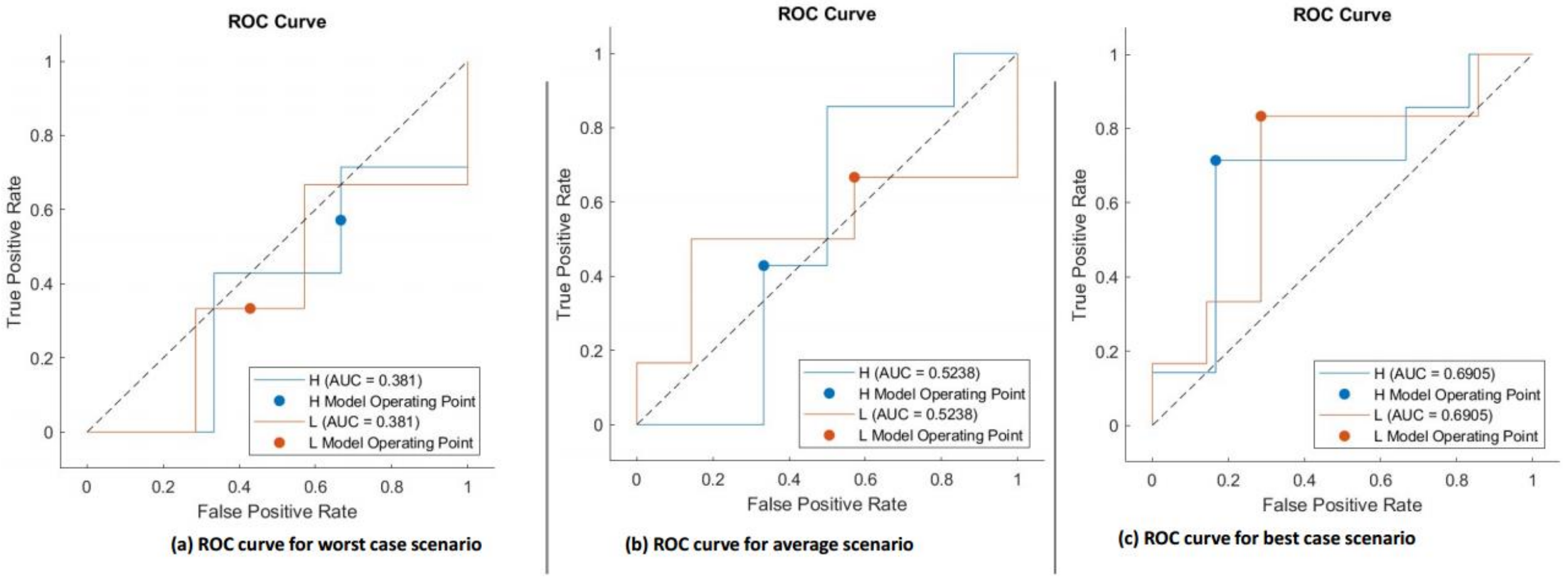

***Figure 9:*** *Performance of the system using Protocol I. During this protocol, the system is tested using all features from all signals. (a,b) Worst and average case scenario where the area under the curve (AUC) varies between 0.38-0.69. (c) A significant increase can be observed in AUC for the best case scenario, where H and L represent healthy and lame cows, respectively.*

***Table 3:*** *System performance for Protocol I. Precision (Pre.), Sensitivity (Sen.), Specificity (Spe.), and Accuracy (Acc.) vary between 46-77% for worst to best case scenarios, where H and L represent healthy and lame cows, respectively.*

| | H | L | Avg. | H | L | Avg. | H | L | Avg. |
|---|---|---|---|---|---|---|---|---|---|
| TP | 4 | 2 | 3 | 3 | 4 | 3.5 | 5 | 5 | 5 |
| FP | 4 | 3 | 3.5 | 2 | 4 | 3 | 1 | 2 | 1.5 |
| FN | 3 | 4 | 3.5 | 4 | 2 | 3 | 2 | 1 | 1.5 |
| TN | 2 | 4 | 3 | 4 | 3 | 3.5 | 5 | 5 | 5 |
| Pre. | 50 | 40 | 46 | 60 | 50 | 54 | 83 | 71 | 77 |
| Sen. | 57 | 33 | 46 | 43 | 67 | 54 | 71 | 83 | 77 |
| Spe. | 33 | 57 | 46 | 67 | 43 | 54 | 83 | 71 | 77 |
| Acc. | 46 | 46 | 46 | 54 | 54 | 54 | 77 | 77 | 77 |

*Protocol II.* Protocol II assesses the most discriminating signals and features (*Table 4)*. From the results, it can be inferred that the angular velocity represents the most discriminating signals (*Table 4-a*). Similarly, the performance of CZT and CDF features is similar as they represent the majority of features used for classification ***(Table 4-b).***

The data distribution used during the study was 70/30%, where 70% represents the training data and 30% represents the test data. Moreover, for all protocols, 10 fold cross-validation was used along with the area under the curve to compare the system performance. The performance is given in Tables 3 and 4 and in Figure 9. The tables and figures show that the selected features can accurately classify the given signal.

***Table 4****: Results of ablation studies in terms of selecting the most discriminating signals and features. Results from Table-a show that angular velocity signals are the most discriminating ones. H, L, and Avg represent healthy, lame, and average cows, respectively.*

*(a) Ablation study for selection of the most discriminating signals*

| | Acceleration | | | Gravity signals | | | Angular position | | | Angular velocity | | |
|---|---|---|---|---|---|---|---|---|---|---|---|---|
| | H | L | Avg. | H | L | Avg. | H | L | Avg. | H | L | Avg. |
| TP | 1 | 6 | 3.5 | 0 | 5 | 2.5 | 3 | 3 | 3 | 6 | 2 | 4 |
| FP | 0 | 6 | 3 | 1 | 7 | 4 | 3 | 4 | 3.5 | 4 | 1 | 2.5 |
| FN | 6 | 0 | 3 | 7 | 1 | 4 | 4 | 3 | 3.5 | 1 | 4 | 2.5 |
| TN | 6 | 1 | 3.5 | 5 | 0 | 2.5 | 3 | 3 | 3 | 2 | 6 | 4 |
| Pre. | 100 | 50 | 54 | 0 | 42 | 38 | 50 | 43 | 46 | 60 | 67 | 62 |
| Sen. | 14 | 100 | 54 | 0 | 83 | 38 | 43 | 50 | 46 | 86 | 33 | 62 |
| Spe. | 100 | 14 | 54 | 83 | 0 | 38 | 50 | 43 | 46 | 33 | 86 | 62 |
| Acc. | 54 | 54 | 54 | 38 | 38 | 38 | 46 | 46 | 46 | 62 | 62 | 62 |

*(b) Ablation studies for CZT and CDF features*

| | CZT Features | | | | | | | | | CDF Features | | | | | | | | |
|---|---|---|---|---|---|---|---|---|---|---|---|---|---|---|---|---|---|---|
| | H | L | Avg | H | L | Avg | H | L | Avg | H | L | Avg | H | L | Avg | H | L | Avg |
| TP | 3 | 2 | 2.5 | 3 | 3 | 3 | 4 | 5 | 4.5 | 3 | 2 | 2.5 | 4 | 3 | 3.5 | 6 | 2 | 4 |
| FP | 4 | 4 | 4 | 3 | 4 | 3.5 | 1 | 3 | 2 | 4 | 4 | 4 | 3 | 3 | 3 | 4 | 1 | 2.5 |
| FN | 4 | 4 | 4 | 4 | 3 | 3.5 | 3 | 1 | 2 | 4 | 4 | 4 | 3 | 3 | 3 | 1 | 4 | 2.5 |
| TN | 2 | 3 | 2.5 | 3 | 3 | 3 | 5 | 4 | 4.5 | 2 | 3 | 2.5 | 3 | 4 | 3.5 | 2 | 6 | 4 |
| Pre. | 43 | 33 | 38 | 50 | 43 | 46 | 80 | 63 | 69 | 43 | 33 | 38 | 57 | 50 | 54 | 60 | 67 | 62 |
| Sen. | 43 | 33 | 38 | 43 | 50 | 46 | 57 | 83 | 69 | 40 | 33 | 38 | 57 | 50 | 54 | 86 | 33 | 62 |
| Spe. | 33 | 43 | 38 | 50 | 43 | 46 | 83 | 57 | 69 | 33 | 43 | 38 | 50 | 57 | 54 | 33 | 86 | 62 |
| Acc | 38 | 38 | 38 | 46 | 46 | 46 | 69 | 69 | 69 | 38 | 38 | 38 | 54 | 54 | 54 | 62 | 62 | 62 |

## 4. DISCUSSION

Discussion section for the present study is extensive as it encompasses the general information about major contributions, future recommendations and comparison with the state-of-the-art. Major contributions and recommendations are covered in public sharing of dataset, signal analysis, effect of machine learning algorithms and development of hand-held devices for lameness detection. Finaly a brief comparison with state-of-the-art is mentioned.

- ***Public sharing of the data***

From the studies mentioned in the introduction (Section 1) and in Table A (Annexture A, Supplementary material), we can see that various sensor-based techniques are being used to predict lameness. However, none of the studies mentioned share data in the public domain, and any referenced data is either commercially available or private in nature (Chapinal and Tucker, 2012). To the best of our knowledge, the present study is the only one sharing a dataset for public research and development. Another unique feature of the study is sharing complete information regarding the hardware and firmware utilized. In Section 2, we have put forth the information on hardware used, which is composed of an Apple iwatch 6, Apple iPhone, and iCloud. Although we have utilized this specific equipment, other combination of smart watch, smart phone and cloud service can be utilized. The only proprietary part in our study is the App which is used to convert data from smart watches to an smart phone. The detailed information just mentioned can be utilized for future development of such projects which are for lameness detection while using sensor-based methodology.

- ***Signal Analysis***

The data we shared is composed of four different types of physical signals, which can be processed individually as reflected in *Table 1*. To further stress the individual characteristics of these signals, an example of data is plotted in *Figure 10*. From the Figure, we can see that acceleration and gravity signals have more dispersion in their values than the gravity signal. Similarly, roll, yaw, and pitch are more reflecting of digital behavior of the magnetometer. Moreover, the morphologies of the signals during the very low amplitudes, as reflected by points 'A' are very different. With the information just presented, the assessment of lameness can be enhanced.

Furthermore, information from similar properties could be utilized for the binary as well multi-class assessment of the lameness. Another aspect of the dataset is the average duration of observation of approximately 6.7 hours, which is helpful to assess the individual motion of a cow. Hence utilizing our dataset, alongside the observation study in a case-control scenario of healthy versus lame cow, we can extend the analysis to identify a cow.

- ***Performance enhancement of machine learning algorithm***

During the Section 1, it was highlighted that lameness is due to multiple causes. Therefore to improve the average performance, new features should be added. Moreover, the information extracted from these features can be enhanced by calculating features at high resolution i.e., instead of using features composed of 370 points, they could be calculated for 1k or even 2k resolution. Features from multiple domains could be utilized to enrich this information. For example, we can extract the lower and upper envelope of a signal using analytical techniques like Hilbert transform. This extraction of envelopes represents the time-based analysis. Moreover, time-frequency decomposition could be utilized to get information at multiple levels. In this regard, techniques like Wavelet, empirical mode decomposition, and others are useful.

The techniques just mentioned can be further divided into goal-oriented and data-driven techniques. We can also utilize numerical techniques like principal component analysis, non-negative factorization, and singular vector decomposition (SVD). All the techniques can be used individually or can be combined. For example, we can apply SVD after decomposing the signal using Wavelets. However, the main purpose is to extract and enrich information useful for detecting lameness. Although the machine leaning technique could be further enhanced as mentioned. However, these techniques lag sequential information that is present in the signal. In this regard, deep learning techniques could be explored to use this sequential information.

Although, we have shared the details of binary classification in Section 3, Results. However, database also offers the possibility of multi-class classification as mentioned in the section of benchmark (Section 2.3) to the research community. Based the framework that was utilized for binary class, multi-class classification achieved 0.21, 0.20, 0.80 and 0.26 in terms of precision, sensitivity, specificity and accuracy in our case. The results show that data has variance which can be utilized but it need additional features with high resolution features and confirms the proposals mentioned above.

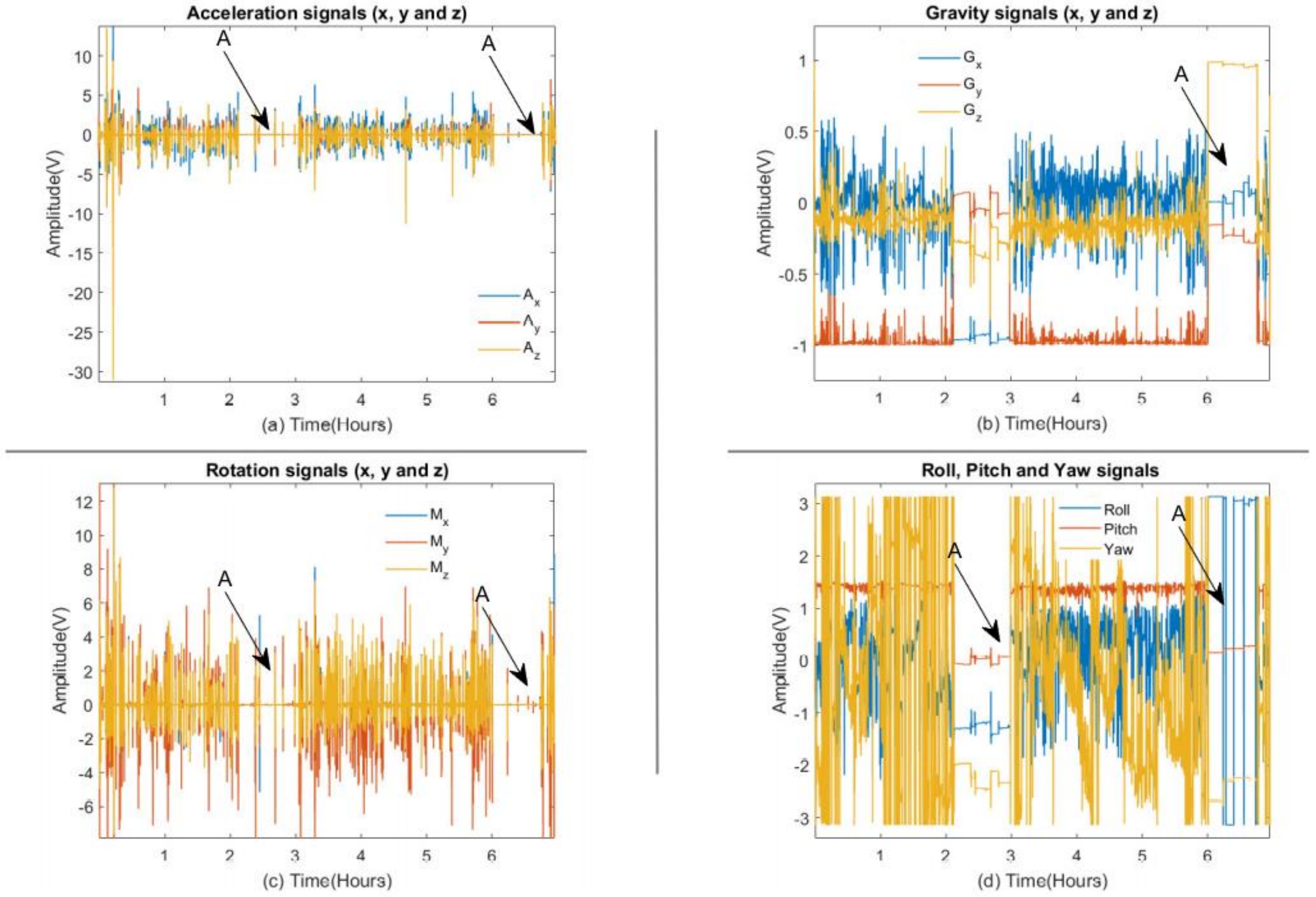

***Figure 10****: Details of the signals with in a multi-sensor data for a cow. (a) Figure shows acceleration signals oriented along x, y and z coordinates. (b) Gravity signals are also oriented along x, y and z coordinates. (c-d) Figures show two types of rotational signals i.e. angular rotation along x, y, z coordinates and roll, pitch, and yaw signals. The arrow 'A' points to the duration of the signals where activity is at minimum.*

- ***Development of hand-held devices for lameness detection***

During this study, we have also shared our base system for binary classification to allow objective comparisons with future studies. This base system utilizes support vector machine (SVM) as a machine learning technique, which is a supervision-based strategy. However, other classification techniques like clustering and Monte Carlo can also be explored. Clustering is an example of an unsupervision based technique and Monte Carlo represents reinforcement learning. The major difference between the supervisory and non-supervisory approaches is the availability of labeled data for classification as supervision-based techniques need initial training before being used during evaluation. Similarly, we can also use semi-supervision-based methods. Another perspective for exploration could be hardware implementation in real-time processing. It represents the transfer of such algorithms to controller-based hardware like Raspberry pi (Raspberry pi, 2023), Atmega and Alf and Vergart's Risc (AVR Microchip Technology, 2023).

- ***Comparison with existing studies***

As highlighted in Introduction (Section 1), objective comparison is not possible with the existing studies due the variation in data (number of files, sampling frequency, duration, etc.). Moreover, we have also highlighted in Data Sampling (Section 2.1.4) that different hardware could be utilized which would still remain a fare comparison as the framework would remain similar. For fair and objective comparison, we are also providing a base algorithm which could be used for the said purpose in future studies. It should also be pointed out that the data provided (43x11,518 = 495,274) can be used for development of machine as well as deep learning based lameness detection as similar, even less, data volumes have already been utilized in studies (Borghart et al., 2021(164*3799 = 623,036); Jarchi et al., 2021(23*25624 = 589,352); Van De Gucht et al., 2017 (45*1240 = 55,800); Van Hertem et al., 2016 (242*3629 = 878,218); Thorup et al, 2015 (348*959 = 333,732); Kokin et al., 2014 (33*481 = 15,873) and Van Hertem et al, 2014 (186*744 = 138,384)). Hence, our study is a breakthrough in the field of lameness detection using inertial data.

The raw data from sensors could be used in multiple ways, like it could predict lameness based on the activity of a cow. As shown in Figure 8, Point A reflects the part of signal where activity of cow is at minimal and these areas could reflect resting position/non activity. The relationship between activity and resting position could be explored. Moreover, the data is being recorded during daily activity as sensor is attached during morning when cow is in milking parlor and is removed in the second session which is generally after 5-7 hours. Hence, the daily life of a cow is recorded which can also be used for transition of health level from healthy to lame.

## 5. CONCLUSION

Our study proves that different sensors in the close proximity can be embedded to create a multi-sensor which is light in weight and is able to record for prolonged period of time. For dairies, the data is useful for lameness detection as we have successfully correlated the lameness with recorded inertial data. We also have hypothesized that this type of sampling can be useful for other quadrupeds like goats, porks, cammels, horse etc. because of similarity of cows' structure with these animal.

We have eliminated the limitation of public access to lameness data by the introduction of CowScreeningDB. Moreover, we have put forth a base algorithm along with the benchmark for objective comparison. Furthermore, a number of analysis techniques are proposed in discussion section from which deep learning, which considers the sequential nature of data, is undoubtly most attractive.

## 5. Acknowledgments

This research was supported by the Spanish MINECO (PID2019-109099RB-C41 project), the CajaCanaria and the la Caixa bank grant, 2019SP19. The authors would like to thank Eva Soler, a veterinarian from the Company of agricultural products of Gran Canaria, Spain, for her assistance during the data collection from the cows.

## Credit authorship contribution statement

The authors would like to extend thanks to BoJinag, Xuqiang Yin and Huaibo Song for their relative contributions as given below:-

Bo Jiang: conceptualization, methodology, software, writing - original draft, visualization, formal analysis, validation.

Xuqiang Yin: data curation.

Huaibo Song: writing - review & editing, supervision, project administration, funding acquisition.

## Declaration of Competing Interest

The authors declare that they have no known competing financial interests or personal relationships that could have appeared to influence the work reported in this paper.

# Appendix A

The appendix presents the literature review and it provides an insight into the types of sensors utilized for lameness detection. It also gives an insight into the public sharing of sensory and non sensory data.

***Table A:*** *Literature review, presented in the form of sensors used, lameness prediction, data statistics, lameness level and public sharing. Data statistics are reflected by sensor, samples from sensors (*) and video/image (!) based studies.*

| Study | Sensor(s) used | Lameness prediction | Data Statistics | Lameness level | Public sharing |
|---|---|---|---|---|---|
| **CowScreeningDB** | **Apple Watch 6** | **Automatic prediction based on raw multi-sensor data** | **43 (12x11.94x24 (11,518))** | **1-5** | **Public** |
| Lemmens et al., 2023 | SenseHub and Landeskontrollverb | Mild lameness detection using combined data of automated milking system and sensor | 374 (31x2682)* | 1-5 | Private |
| Jiang et al., 2022 | Video from PAL (phase alteration line) camera | Automatic cow lameness detection from video using cow's back position extraction and cow's back curvature extraction models | 90 (810x1920x1080 x40s)!! | 1-3 | Private |
| Frondelius et al., 2022 | IceQube | Automatic lameness detection using predictors of lying time, number of lying bouts, maximum length of the lying bout, roughage feeding time, number of visits to the feeder and number of steps | 85 (6x266x24) | 0-7 | Private |
| Borghart et al., 2021 | MooMonitor and Dairymaste | Prediction of lameness using 3D acceleration, milk based predictors, resting, feeding, grazing and weight distribution | 164 (6x3799)* | 1-5 | Private |
| Jarchi et al., 2021 | Bosch BMI160 Inertial measurement unit | Lameness detection by measuring instantaneous frequency using data from inertial measurement unit | 23 (8x25624)* | 0-3 | Private |
| Antanaitis et al., 2021 | RumiWatch noseband halter | Association of lameness with rumination behavior by using noseband activity | 20 (3x13x24) | 1-5 | Private |
| Riaboff et al, 2021 | RF-Track and Video camera | Automatic prediction based on acceleration & location data using videography to record lameness predictors of grazing & resting behaviors and the position | 68 (5x35x24) | 0-3 | Private |
| Shahinfar et al, 2021 | - | Lameness prediction based on general features and production, linear & composite type traits | - (29x2535) | 0-3 | Private |
| Kang et al., 2020 | Video Recording | Lameness detection based on supporting phase (Time between lifting and landing time of a hoof) | 600 (1000x1,920 × 1,080)! | 1-3 | Private |
| Taneja et al., 2020 | Long-Range Pedometer, ENGS | Automatic lameness detection based on step count, lying time and swaps. | 150(3x-) | 1-4 | Private |
| Jiang et al., 2020 | Webcam and SONY HDR-CX290E | Automatic cow lameness detection from real time video | 1080 (756*x1920x108 0x40s)!! | 1-4 | Private |
| Piette et al., 2020 | 3D Camera | Real time cow lameness detection using back posture and window length ( Historical reference for comparison for lameness) | 209 (925x42x24) | 1-5 | Commercial |
| Byabazaire et al., 2019 | Long-Range Pedometer, ENGS Systems | Early detection of lameness using online information sharing | 146 (3x12x24) | 0-1 | Private |
| Jiang et al., 2019b | DS-2DM1-714 dome cameras | Automatic cow lameness detection from real time video | 30 (360x704x576x3 0s)!! | 1-2 | Private |
| Weigele et al., 2018 | MSR145 data Logger and RumiWatch halter | Moderate lameness detection using 3D acceleration, laying behavior, locomotors activity, neck and nose activities | 208 (13x2x24) | 1-5 | Private |
| Barker et al., 2018 | Omnisense Series 500 Cluster Geolocation System | Combining local positioning and acceleration measurements for feeding behavior to detect lameness | 19 (5x2x20m) | 0, 2 | Private |
| Vázquez Diosdado et al., 2018, | Ominsense Series 500 Cluster Geolocation System | Relationship between lameness and cow mobility | 210 (3x5x24 (900,000)) | 0-2 | Private |
| Van De Gucht et al., 2017 | Gaitwise system | Automatic cow lameness detection with a pressure sensor | 45(1240)* | 1-3 | Private |
| Beer et al., 2016 | HERDE 5.8, herd management program, Nikon Coolpix L830 Two 3D RumiWatch and noseband sensor | Automatic detection using multimodal data which is composed of acceleration, motion detection and video processing using eating and ruminating time, eating chews & ruminating chews, lying time and lying bout duration, lower standing time. | 53 (18x11x24) | 0-1 | Private |
| Thorup et al., 2016 | Silent Herdsman accelerometer and RumiWatch halter | Establish relationship between lameness and rumination predictors (rumination time, number of rumination events, feeding time, feeding frequency, feeding rate, feed intake, and milk yield) | 16 (17x19.4x24) | 0-1 | Private |
| Van Hertem et al., 2016 | DeLaval sensors and KinectTM | Image processing using posture, behavior and performance parameters of 3D depth, milking order and heart detection | 242 (3x3629)* | 1-5 | Private |
| Thorup et al, 2015 | IceTag3D | Acceleration data composed of daily lying duration, standing duration, walking duration, total number of steps, step frequency, motion index. | 348 (3x959)* | 1-4 | Private |
| Kokin et al., 2014 | IceTag3D and video recording | Synchronized data analysis of acceleration and videography composed of parameters such as the number of steps, motion index, lying bouts, standing and lying time | 33 (5x481)* | 1-5 | Private |
| Garcia et al., 2014 | DeLaval activity tag | Lameness detection in automated milking systems using partial least squares discriminant analysis | 88 (320x70x24) | 1,3,4 | Private |
| Van Hertem et al, 2014 | Video recording using Microsoft Kinect Xbox 3D-camera | Automatic lameness detection using consecutive measurements of an individual cow | 186 (3x744)* | 1-5 | Private |
| De Mol et al., 2013 | IceQubes sensors | Lameness alert generation based on correlation between 3D accelerator and milking data | 100 (9x64,040)* | 1-3 | Private |
| Van Hertem et al., 2013 | FreeFlow and HR-Tag | Enhancement of lameness detection by using milk yield, neck activity and ruminating time based sensors in presence of heat detection and monitoring of the nutritional status of the herd | 118 (4x56x24) | 1-3 | Private |
| Kamphuis et al., 2013 | Waikato milking systems and pedometer | Automatic detection based on weight, milk measurement and activity | 4904 (11x14x24) | | Private |
| Chapinal and Tucker, 2012 | Pacific industrial scale and DCRSR100 | Automatic detection using weight and gait score | 63 (2x-x8m) | 1-5 | Private |
| Maertens et al., 2011 | Gaitwise system | Real time cow gait tracking and analyzing tool to assess lameness using a pressure sensitive walkway | 174 (20x70x24) | 1-3 | Private |
| Nielsen et al., 2010 | IceTag3D and HDR-SR10E | Automation detection using motion leg analysis | 10 (3x177x10m) | - | Private |
| Pastell et al., 2008 | Electromechanical film sensor | Automatic detection using weight and gait score | 43(2x-) | 1-5 | Private |
| Pastell et al., 2006 | Piezo electric force sensors | Lameness measurement using force measurement | 50(-) | - | Private |